# LAND: A Longitudinal Analysis of Neuromorphic Datasets

Gregory Cohen
*International Centre for Neuromorphic Systems (ICNS)*
*Western Sydney University*
Sydney, Australia
g.cohen@westernsydney.edu.au

Alexandre Marcireau
*International Centre for Neuromorphic Systems (ICNS)*
*The University of Manchester*
Manchester, United Kingdom
alexandre.marcireau@manchester.ac.uk

*Abstract*—Neuromorphic engineering has a data problem. Despite the meteoric rise in the number of neuromorphic datasets published over the past ten years, the conclusion of a significant portion of neuromorphic research papers still states that there is a need for yet more data and even larger datasets. Whilst this need is driven in part by the sheer volume of data required by modern deep learning approaches, it is also fuelled by the current state of the available neuromorphic datasets and the difficulties in finding them, understanding their purpose, and determining the nature of their underlying task. This is further compounded by practical difficulties in downloading and using these datasets. This review starts by capturing a snapshot of the existing neuromorphic datasets, covering over 423 datasets, and then explores the nature of their tasks and the underlying structure of the presented data. Analysing these datasets shows the difficulties arising from their size, the lack of standardisation, and difficulties in accessing the actual data. This paper also highlights the growth in the size of individual datasets and the complexities involved in working with the data. However, a more important concern is the rise of synthetic datasets, created by either simulation or video-to-events methods. This review explores the benefits of simulated data for testing existing algorithms and applications, highlighting the potential pitfalls for exploring new applications of neuromorphic technologies. This review also introduces the concepts of meta-datasets, created from existing datasets, as a way of both reducing the need for more data, and to remove potential bias arising from defining both the dataset and the task. Finally, this paper and the accompanying List of Available Neuromorphic Datasets (LAND) tool, provide a comprehensive catalogue of datasets to allow researchers to effectively locate relevant data before attempting to collect new datasets.

## I. Introduction

In this paper, we present the outcomes of a longitudinal analysis into neuromorphic datasets and provide insight into the current relationship between the neuromorphic community and their data. We examine 423 datasets, from 386 different academic publications, totalling over 41 TB of data.

This review delves into the nature of the available datasets and examines the trends in their generation, purpose, and distribution methods. Through this analysis, we identify important considerations around the nature of neuromorphic datasets, especially in comparison to data obtained from conventional sensor systems. Based on these insights, we identify issues in how the community produces and publishes data, and provide recommendations for mitigating these issues.

As the vast majority of datasets contain vision-based neuromorphic data (412 out of 423), the analysis of the data in this paper is heavily skewed toward vision, however, care has been taken to include non-vision datasets and to ensure that the conclusions and recommendations of this work also apply to other sensor modalities.

### A. Datasets: Their importance, relevance, and complexities

The focus and purpose of this paper is to explore the interactions between the community and their data. Whilst the need for datasets underpins these interactions, we will only briefly cover the motivations for producing, sharing, and using datasets. There is little need to motivate for the importance of datasets - as data, and by extension datasets, have transcended from an important part of research to become the basis for entire research fields. Often the data drives the research, and neuromorphic engineering is experiencing a similar shift toward a data-driven research approach.

There has been acknowledgement of the importance of neuromorphic datasets for about as long as there have been neuromorphic datasets to discuss. The role of data in benchmarking neuromorphic vision systems was a topic of debate and discussion as early as 2015 [1], with researchers already highlighting the pitfalls of direct comparisons to conventional frame-based datasets. The suitability of conventional datasets for use in neuromorphic systems was questioned as early as 2015, with closed-loop benchmarking tasks proposed as a better, albeit more challenging means of assessing neuromorphic systems [2]. Our lab echoed this sentiment, proffering that closed-loop systems are the gold-standard for comparing neuromorphic and conventional systems [3], yet offering little in the way of practical means by which to produce them. More recently, the Neurobench framework [4], has sought to provide a open-source platform for the development and assessment of community-created benchmarks.

The motivations for creating neuromorphic datasets usually focus on lowering the barrier to entry for the field by removing the need for expensive neuromorphic hardware, reducing expensive and time-intensive data collection campaigns, allowing researchers to focus on the algorithms and hardware over data collection, and to jumpstart the field through large-scale datasets in the same manner as the fields of computer vision and deep learning. Most importantly, they provide a means of comparing algorithms and measuring and charting success and advancements in the field.

In this paper, we hope to provide the community with data-driven guidance on the current state of neuromorphic datasets and to provide tools, guidelines, and observations to ensure that the field creates meaningful, long-lasting, and representative datasets.

### B. What makes a dataset neuromorphic?

Defining exactly what constitutes neuromorphic engineering is a complicated task and it is unlikely that any defining guidelines will reach a consensus within the community. Our criteria for determining whether a dataset is neuromorphic draws heavily from the definitions of neuromorphic sensory systems in [5]. Therefore we consider most datasets that broadly follow the approach described in that paper to be neuromorphic.

As a general rule of thumb, any data that is collected with a neuromorphic sensor, such as an event camera or a silicon cochlea, are inherently neuromorphic. Data that is provided in the Address-Event Representation (AER) [6] is also almost always considered to be a neuromorphic dataset. This can include data from tactile sensors, for example, where the device may not operate asynchronously but still emits data in the AER format.

Datasets that consist of biological recordings of neurons are not generally considered to constitute a neuromorphic dataset, unless the intended purpose of the data is to build, test, or characterise a neuromorphic system or algorithm.

### C. What constitutes a dataset?

For the purposes of this paper, we define a dataset to be an organised collection of data used in a research context. Given that most papers in neuromorphic engineering present an algorithm or data interpretation technique, the majority of papers will involve the processing of data. To distinguish between data that is solely used to validate an algorithm and a collection of data intended to be a dataset, we have compiled a set of test criteria. For the purposes of this work, a collection of data is considered to be a dataset if:

1) The associated paper refers to the data as a dataset, or compares the data to other existing datasets.

2) The data was deliberately collected or collated with a particular structure or organisation.

3) The data or method of producing the data is made publicly available, or intended to be made public.

4) There is a clear, well-defined task associated with the data and a comparable metric of performance.

Data that is collected from a system solely for the purposes of showcasing performance is usually not considered to constitute a dataset. For the purposes of this paper, the DVS09 dataset [7], which contains an assortment of event-based data recordings is included, despite not having a clearly defined task, as the data has been widely used in other publications.

Additionally, some datasets do not contain new data, but instead build upon existing datasets by providing additional annotations or labelling. For example, the DSEC-MOT [8] dataset provides a multi-object tracking task through a set of extended annotations for the DSEC dataset [9]. A further example is the DVS Gesture-C dataset [10], which is a modified version of the DVS Gesture dataset [11] in which different types of noise are mixed into the data to test robustness of classification algorithms. Koizumi and Watanabe created an augmented version of the DHP19 [12] dataset in which the subjects are scaled and translated through the field of view to create a new annotated version of the dataset for localisation and size estimation [13]. These are considered to be independent datasets for the purposes of this work, although the data that they contribute is only counted if it is modified and distributed separately from the original dataset.

## II. THE ISSUE OF REUSABILITY

There are two primary purposes of a dataset: reproducibility and reusability. Both of these are crucially important in scientific research. The need for reproducibility in research has never been more prevalent, given the rise of generative AI tools used at every stage of the research process. This is likely to become a more increasing necessity as the adoption of neuromorphic engineering continues to grow.

On the other hand, the importance of reusability has far reaching effects, as it enables future research, reduces the need to collect more data, and provides means of benchmarking and charting research progress. This is where the neuromorphic community finds itself in trouble.

### A. Methodology for exploring dataset re-usability.

There are few direct means of measuring the usage of datasets. Although some data repositories track downloads and even host leader boards for performance, these platforms only host a small fraction of the available datasets - a topic more thoroughly discussed in Section III. However, most datasets are introduced through an academic publication (386 out of 423 datasets), and we use the citations for those papers as a proxy for measuring its usage and impact.

This is not a perfect metric, as papers may be cited in cases where the data itself was not used. However, the reference still indicates that the dataset was used in the research process. There are also papers in which the introduction of the dataset is secondary to the main contribution of the work, and these papers may get cited separately to the underlying data. There are also datasets that are introduced without an accompanying academic paper, although these constitute only a small portion of the available datasets.

Data for this analysis was sourced from two primary sources. As the underlying collection of datasets was compiled, each dataset was assigned to an academic paper wherever possible. Details on these papers were compiled from publicly available sources such as CrossRef and Google Scholar. This data was then augmented by commercial citation data from Lens.org[1], which provided a far more comprehensive list of citing and cited papers for our analysis. Whilst we have made an effort to publicly release all the data used in this study, some of the citation information can only be made available on request.

---

[1] https://www.lens.org/

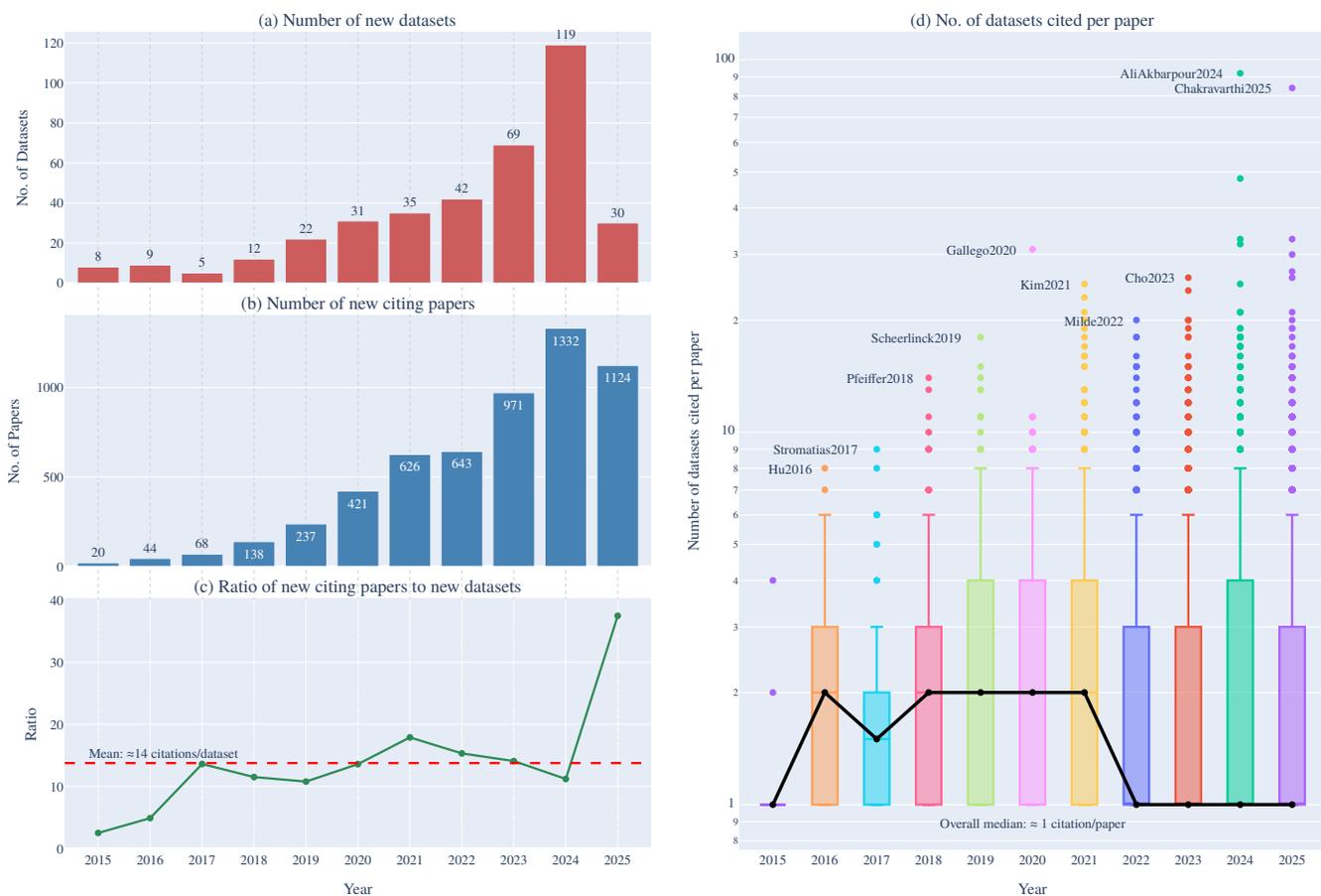

Fig. 1. **The growth in datasets, citing papers, and cited papers in neuromorphic research.** The above figures explore whether the community is indeed re-using datasets. The growth in the number of available datasets is shown in (a), which has risen sharply over the past five years. Alongside this rise, the number of papers citing neuromorphic datasets has also risen sharply. as shown in (b). The ratio of these two yearly values, as shown in (c), indicates that approximately 14 datasets are cited for each new dataset released. Whilst this appears to show a healthy research community, digging into those statistics reveals a more complicated truth. When examining the number of datasets cited per paper, as shown in (d), it is clear that the majority of the yearly citations come from large review papers (with several of the most widely citing papers labelled). In fact, the median number of datasets cited per paper, emphasised by the black line drawn over (d), over the 10 years of this study is almost exactly equal to 1, indicating that most papers will cite only a single dataset.

### B. The Good: More datasets and more citing papers

This is perhaps the most important question posed in this paper - does the field truly re-use neuromorphic datasets? The answer is, somewhat unsurprisingly, that it is complicated. This section will explore that issue in depth, starting with Fig. 1, which shows a series of charts drawn from the citation data collected in this study.

Perhaps it is best to start with the positive aspects. The top-left panel in Fig. 1 shows the number of new datasets published each year. Note that there was only one officially published dataset in 2014, which is why we have elected to only plot data from 2015 onwards. It is clear from this figure that the interest in neuromorphic data has grown spectacularly since 2021, with an almost exponential increase in the number of available neuromorphic datasets. There are now at least 423 publicly available datasets for the community.

Following on from this, and turning our attention to Fig. 1 (b), which shows the number of papers citing neuromorphic datasets per year. Each paper that cites a neuromorphic dataset in this collection is only counted once, regardless of the number of datasets cited. As with the number of datasets, there is a striking and correlated growth in the number of papers that cite a neuromorphic dataset.

It is interesting to note the significant decrease in the number of datasets published in 2025. This may simply be an artefact of the collection process for this data, as data collection for this study ended in the last quarter of 2025. However, it may also be an indication in a shift away from data collection and toward data processing, potentially driven by a strong interest in data-hungry deep learning techniques. Whilst it is too early to lend any credence to that theory, it is supported by the far less significant decrease in the number of citing papers in 2025. This seems to indicate that there is no waning interest in the field, but perhaps a waning interest in collecting new data.

Calculating the ratio of the number of new datasets to the number of papers citing datasets yields the ratio shown in Fig. 1 (c). This plot shows both the yearly ratio and a horizontal line indicating the mean over the 10 years of the study. Computing this mean indicates that there are an average of 14 citing papers to each new dataset released, and this has stayed relatively consistent over the past ten years.

TABLE I
List of papers that cite the most datasets

| Cites | Year | Paper Title |
|---|---|---|
| 92 | 2024 | Emerging Trends and Applications of Neuromorphic Dynamic Vision Sensors: A Survey [14] |
| 84 | 2025 | Recent Event Camera Innovations: A Survey [15] |
| 48 | 2024 | Event Cameras in Automotive Sensing: A Review [16] |
| 33 | 2024 | Brain-Inspired Computing: A Systematic Survey and Future Trends [17] |
| 33 | 2025 | Hardware, Algorithms, and Applications of the Neuromorphic Vision Sensor: A Review [18] |
| 32 | 2024 | An Application-Driven Survey on Event-Based Neuromorphic Computer Vision [19] |
| 31 | 2020 | Event-based Vision: A Survey [20] |
| 30 | 2025 | Event-Based Visual Simultaneous Localization and Mapping (EVSLAM) Techniques: State of the Art and Future Directions [21] |
| 27 | 2025 | EventAid: Benchmarking Event-Aided Image/Video Enhancement Algorithms With Real-Captured Hybrid Dataset [22] |
| 26 | 2023 | Label-Free Event-based Object Recognition via Joint Learning with Image Reconstruction from Events [23] |
| 26 | 2025 | NER-Net+: Seeing Motion at Nighttime With an Event Camera [24] |
| 25 | 2024 | Recent Advances in Bio-Inspired Vision Sensor: A Review [25] |
| 25 | 2021 | N-ImageNet: Towards Robust, Fine-Grained Object Recognition with Event Cameras [26] |

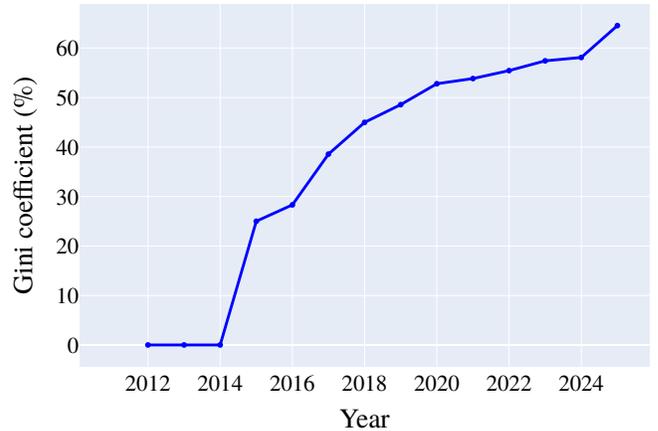

Fig. 2. **Gini coefficient plot showing the inequality in the distribution of dataset citations.** The Gini coefficient is a metric, most commonly used in economics, that measures the inequality in distribution amongst a population. In the above figure, datasets that are at least one year old constitute the population and the measure for which we are exploring inequality is the number of new citations for that year. If every dataset cited every other dataset, this would yield a Gini coefficient of zero, and if all datasets only cited a single dataset, the Gini coefficient would be one. The above graph shows a relatively high Gini coefficient, indicating that there is significant inequality in the distribution of cited datasets, indicating that the most papers only cite a small subset of the available datasets.

*C. The Bad: Dataset papers do not cite other datasets papers*

Whilst the ratio shown in Fig. 1 (c) seems promising, digging further into the data uncovers deeper issues and illustrates why the average number of new citations per new dataset can be misleading.

Fig. 1 (d) shows a box-and-whisker plot showing the number of datasets cited per paper from 2015 to 2025. Outliers are shown as points, and the number of datasets cited per paper is shown on a logarithmic axis. Additionally, the figure has been annotated to show an overlaid plot of the median number of datasets cited per paper using a solid black line. The data clearly shows that the majority of dataset papers published cite only one other neuromorphic dataset paper, and that has been consistent over the past ten years, if not becoming more prevalent after 2022.

From 2021 onward, there is an increased number of papers that appear to cite a significantly higher number of papers. These are clear outliers, often citing an order of magnitude more dataset papers than most of the papers from that year. Whilst promising, this is likely more of an indication of the growth of the field rather than a select few adopting better practices. Table I provides the results of examining these outlying papers and shows the titles of the papers citing the most datasets and their citation counts. It is clear from the list, and perhaps rather unsurprising, that most of these papers are survey and review papers.

*D. The Reality: Inequality in the papers that get cited*

Digging further into the data, we can examine the inequality present in the citations of neuromorphic datasets using tools from classical economics. In this case, we will borrow tools used to measure wealth inequality, which is often measured and reported using a metric known as the Gini coefficient. Derived from the Lorenz curve, it provides a scalar metric of inequality that ranges from 0 (perfect equality) to 1 (perfect inequality).

The Lorenz curve models the cumulative share of a resource (y-axis) held by the bottom percentage of units (x-axis). It is often used to make statements such as "the bottom 50% of the population holds 20% of the wealth". In our case, we construct a Lorenz curve for each year, with the resource being the new citations in that year and the units being the number of datasets that are at least one year in age.

We only consider datasets that have been published for more than a year to account for the delays in publishing of citing articles, thereby allowing these new datasets to gain some citations before being considered. Similarly, for each paper considered, we count each dataset cited as a separate citation.

The Gini coefficient is calculated from the Lorenz curve using the 45° line of equality. It measures the ratio of the area between the Lorenz curve and the line of equality, divided by the total area under the line of equality. This results in a measure that increases as the Lorenz curve diverges from the line of equality, and reduces as it moves closer. For each year, we calculate the Gini coefficient of the new citations against the available datasets to derive a measure of the inequality in the citations across the existing datasets, as shown in Fig. 2.

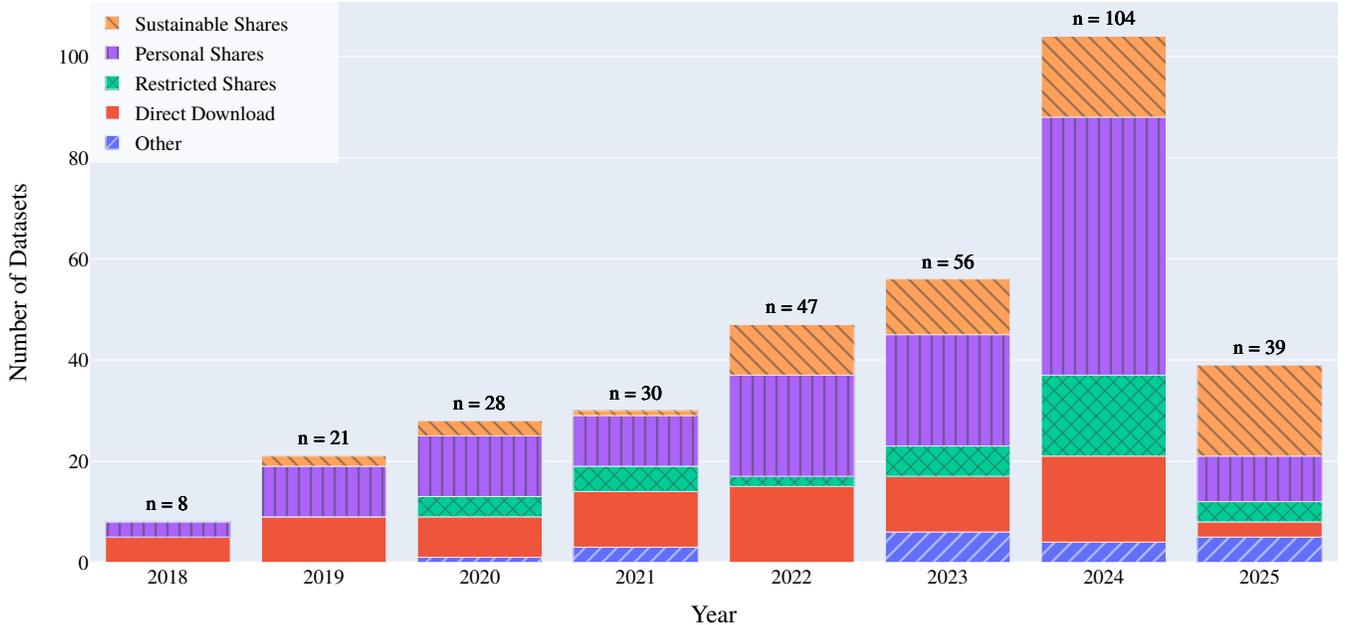

Fig. 3. **Methods of dataset distribution over time.** The above figure shows the methods used to distribute neuromorphic and event-based datasets. These have been categorised into five categories based on the nature of the sharing mechanism. The total number of datasets in each year is provided above each bar. Most notable is the rise in personal sharing platforms that are tied to an individual, and the rise of platforms that place restrictions on users downloading the data.

The plot of the Gini coefficient reveals the level of inequality within the distribution of citations across datasets. Up until 2015, there was only one publicly available dataset, which results in perfect equality and therefore a Gini coefficient of zero. The Gini coefficient quickly rises as more datasets become available and has continued to slowly increase. In 2025, the Gini coefficient reached its highest value of 0.65, perhaps related to the decrease in the number of new datasets created, evident in Fig. 1 (a). Interestingly, the Gini coefficient in 2025 bears similar resemblance to the same coefficient calculated when looking at global citation trends [27].

The fact that we get a fairly high Gini indicates high inequality and therefore that some datasets get a far bigger share of new citations than others. Combined with the results from Fig. 1, this indicates that most papers not only cite one or two datasets, but that they also all tend to cite the same small subset of datasets.

## III. The Issue of Data Availability

To date, the neuromorphic community has generated at least 40.95 TB of neuromorphic data that has been labelled as publicly accessible. However, accessing this data is often difficult and wrought with issues. The challenges of how to share and distribute this data have become an increasingly difficult and complicated problem, and one that the community needs to tackle to drive data re-use over data creation.

In this section, we examine the trends in sharing methods for neuromorphic datasets in the community, spanning the 327 datasets that have identified distribution methods.

### A. Dataset distribution categories

We can categorise the distribution methods for neuromorphic datasets into five different categories:

**Sustainable Shares** are the gold-standard method for storing data. These platforms are designed to support reproducible research, providing long-term data storage and open access to the data. These platforms also make the data visible externally, providing indexing, searching, and citing. Examples of these include Zenodo, IEEE DataPort, FigShare, and HuggingFace.

**Personal Shares** are data hosting services that are linked to an individual, usually through a personal data sharing service such as Google Drive or Microsoft OneDrive. This makes continued access to the data highly dependent on the individual, often remain valid only when the user is active in the research community. These can also be corporate accounts, such as Microsoft SharePoint, which are again tied to an individual and often disappear when the person leaves the organisation.

**Restricted Shares** are data access platforms that implement access restrictions on the data or datasets. These platforms often require a registration that requires a mobile phone number from a specific country, or services that implement geo-based restrictions on data download.

**Direct Download** datasets encompass all the datasets where the data is stored on a publicly accessible server and can be directly downloaded through the browser without any restrictions or limitations. Usually, these are university-hosted servers that provide direct access to the files. The defining characteristic of these shares is that they allow for automated access to the data without being tied to an individual, a file-sharing platform, or rate or access limitations. These methods for sharing data are typically highly reliable.

**Other Shares** include other sharing options that do not fall into the above categories. Examples of these include peer-to-peer file sharing tools, re-purposed file sharing tools, and proprietary systems. Often these methods require a specific application to be installed to provide access to the data.

*B. Trends in data distribution*

Fig. 3 shows an illustration in the trends of data distribution methods since 2018. For each year, the bar shows the breakdown of different distribution methods as a percentage of the datasets released in that year, with the total number of datasets created in that year provided as an annotation above each bar.

Most noticeable in the data is the increasing use of personal file sharing sites to distribute neuromorphic data, currently being used for 42% of all datasets, and 32% of all neuromorphic data available. This is a concerning trend as these links are tied to an individual, meaning that their persistence is inherently tied to that individual. Once an academic leaves the field, or even changes universities, these shares often become inaccessible.

Personal file sharing tools often have hidden usage limits that cause unpredictable issues for access and often prevent automated access to the datasets. These tools also make it difficult to determine the total size of the dataset, and greatly complicate access to datasets comprising lots of small files. Datasets are then compressed into a single monolithic archive to mitigate these issues, often requiring the download of large files simply to determine their contents. The large size of these files also makes them prone to corruption when transferred and more difficult to verify once uploaded.

There is also a noticeable rise in the use of file sharing platforms that have some geophysical restrictions that limit access to the data. These platforms often require the user to provide a mobile phone number from a specific country to register for an account, greatly limiting the access to the data. Websites that are widely accessible but blocked in certain countries are not included in this category, as the restrictions do not originate from the distribution platform itself. This data is therefore inaccessible for use and for reproducibility purposes.

Finally, and encouragingly, is the rise in the use of sustainable data sharing platforms, such as Zenodo and HuggingFace, which are designed to host datasets and make them publicly accessible. These platforms allow for dataset versioning and provide DOIs, allowing for proper referencing.

Table II shows a list of the most cited datasets alongside their distribution methods and total dataset size. It is immediately apparent that the most widely adopted datasets almost all provide direct download options. This is likely due to the ease with which the data can be accessed, and retrieval automated. Several of the datasets, such as The Multivehicle Stereo Event Camera Dataset (MVSEC) [28] and The Driving Scenarios for Event Cameras Dataset (DSEC) [9], have detailed project pages that provide individual links to data files alongside detailed descriptions of their contents and even short video previews.

TABLE II
Distribution methods for the top 10 most cited datasets

| Dataset | No. citations | Size (GB) | Distribution Methods |
|---|---|---|---|
| DVS-GESTURE [11] | 1167 | 2.7 | Direct Download |
| N-Caltech101 and N-MNIST101 [29] | 993 | 3.72 | Google Drive, Dropbox, OneDrive |
| The Event-Based Camera Dataset [30] | 784 | 19.97 | Direct Download |
| Rebecq2019 [31] | 753 | 44.7 | Direct Download |
| CIFAR10-DVS [32] | 630 | 7.81 | FigShare |
| MVSEC [28] | 594 | 112.0 | Direct Download, Google Drive |
| N-CARS [33] | 565 | 0.29 | Direct Download |
| DSEC [9] | 408 | 150.0 | Direct Download |
| EED [34] | 407 | 0.2 | Direct Download |
| Bardow2016 [35] | 386 | N/A[2] | Direct Download |

The only datasets that make use of personal shares are the N-MNIST and N-Caltech101 datasets, which were introduced together in a single paper [29]. These datasets are relatively small, making it easy to distribute as a single monolithic compressed file. Additionally, multiple personal sharing platforms are provided, with the caveat given that some may be unreliable when there are lots of concurrent accesses.

*C. Data access licenses*

Several datasets require the explicit acceptance of license terms prior to granting access to the data. Several datasets, such as ADD [36] and EvPointMesh [37] require an online form to be completed before granting access to the data. Others, such as PKU-Spike-Recon [38] require a form to be emailed to the authors. The EHPT-XC dataset [39] even requires a submission with a handwritten signature. Other datasets, such as EventSleep [40] require registering on a specific platform to download the data. Datasets hosted on IEEE DataPort, such as the The Spiking Heidelberg Digits Dataset (SHD) [41], also require a user account to access the data.

Whilst some license agreements may be necessary, the additional steps required to access the data both reduce the ease of use and also prevent automated access to the data. Datasets can be distributed with license files that dictate the terms of use, and this should be the preferred choice in every situation.

---

[2]Download links for Bardow2016 are currently broken, making it difficult to assess the size of the official dataset.

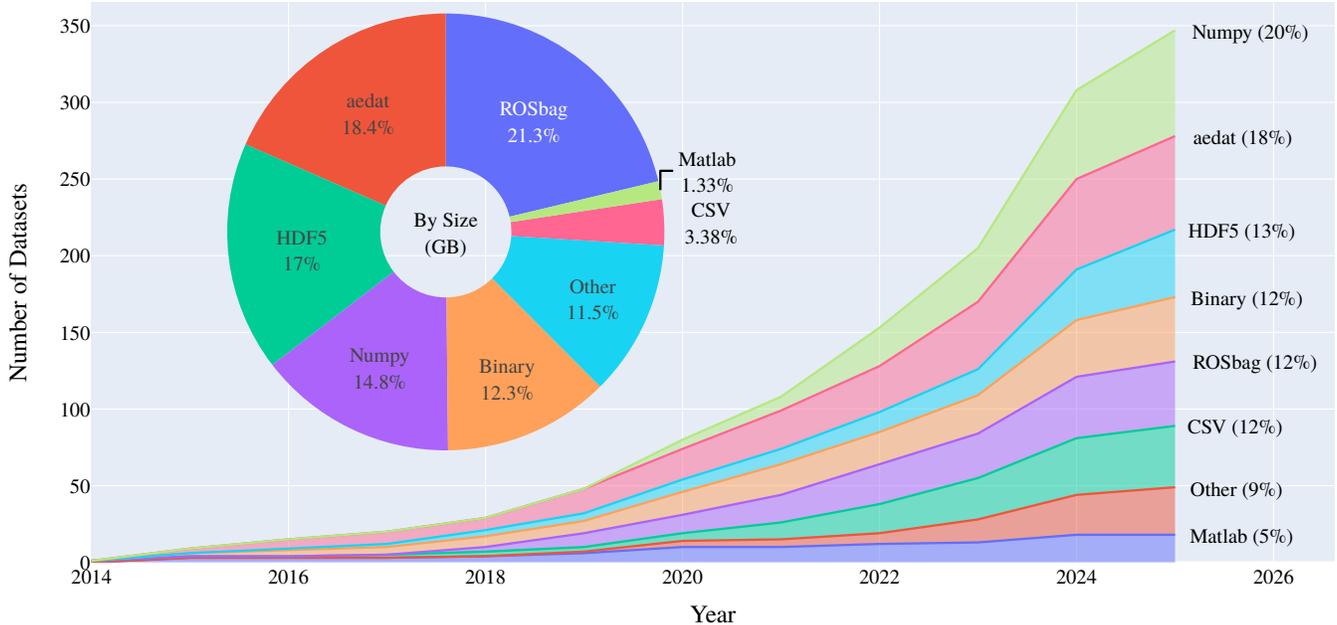

Fig. 4. **The range of dataset formats used in distributing neuromorphic data.** The above figure shows the trends in adoption of different file formats for distributing neuromorphic datasets. The area graph shows the cumulative number of datasets in each file format year on year whilst the inset pie chart shows the breakdown of the use of different file formats in terms of data volume. Whilst there is strong growth in the use of easily accessible formats, such as `Numpy`, there is still hidden complexity in accessing the data in all of the provided formats.

## IV. THE ISSUE OF DATA ACCESSIBILITY

Closely related to the challenges of data availability discussed in Section III, are the issues of data accessibility, which encompass how the data in available datasets can be understood, extracted, and ultimately used. This is a multi-faceted problem, broadly comparable to the issue with differing file formats in conventional datasets, but complicated by the lack of official or community standards for the storage and processing of neuromorphic data. This analysis begins by examining the different file formats and storage mechanisms used across the 337 datasets with known file formats, exploring the trends in the choice of datasets both by the quantity of datasets using the format, and also by the total volume of data stored in each format.

Fig. 4 shows a breakdown of the different file formats across the neuromorphic datasets. The area plot shows the growth in dataset formats year-on-year in terms of the number of datasets using each file format. Inset into the figure is a pie chart showing the total breakdown of file formats by data volume from the 299 datasets with known data sizes.

### A. Neuromorphic file formats

A persistent issue with data formats is the requirement for specialised tools, software, or libraries required to access the data. Further complicating the issue, the prevalence of data distributed in large monolithic archives often require the whole dataset to be downloaded before the file format can be inferred. Even the use of more general file formats, such as text files and `Numpy` files, often still requires context to understand and work with the data.

Text file formats, such as (`txt`, `csv`, and `tsv`) files are easy to explore and work with programmatically. Their downside is that they tend to be very large when uncompressed, and the format of data within these files can vary widely. Two primary issues are the ordering of the values in these files (i.e. $e = [x, y, t, p]$ vs. $e = [t, x, y, p]$), whether integer or floats are used for the $(x, y)$ coordinates, the chosen polarity convention (i.e. $p \in [0, 1]$ or $p \in [-1, 1]$). These differences are not necessarily difficult to overcome, but complicate automation and often need to be determined from the data.

The `HDF5` format is a highly versatile container designed to support multiple datasets in a single file and allows metadata to be associated alongside each dataset. It contains structured data that can be interrogated with tools and programmatically, and is widely supported across programming languages. However, there is very little consistency in the layout and organisation of these files in the neuromorphic community, often compounded by a lack of description or documentation. The native format does provide the ability to compress the data but certain datasets, such as <u>DSEC</u> [9] make use of plug-ins to provide additional compression, which can further complicate adoption. Perhaps the best use of `HDF5` files is to store datasets, such as <u>N-MNIST</u> [29], which is comprised of several thousand small files. However, the dramatic increase in sensor resolutions has resulted in far fewer datasets of this composition.

Matlab data files (`.mat` files) are essentially `HDF5` files internally and provide a convenient means of persisting and loading data into Matlab. These have somewhat fallen out of favour within the community, evident by the early plateauing of their use shown in Fig. 4. As with `HDF5` files, this format does support compression but it struggles to scale with dataset size. As Matlab can natively read `HDF5` files, there is little reason to use `.mat` files over standard `HDF5` files.

`ROSBags`, the data storage system used by Robot Operating System (ROS) is widely used in the neuromorphic community, especially in robotic and vehicular-based datasets. It allows the bundling of multiple sensor types into a single file and with synchronised timing. It is very convenient when working with a ROS-based ecosystem, but requires a large tool chain to access the data, complicating the use of the data on other platforms. `ROSBag` files can also grow very large, especially as they often contain other high-volume data streams, such as conventional video and LiDAR data.

Camera-specific binary files, such as the `aedat` formats and the `EVT` formats, pack the event-based data into a container file format that is often not human-readable. The files therefore contain predictable and structured metadata and support effective compression methods. However, these formats usually require software libraries or tools to access the data.

The tools provided to operate the iniVation range of cameras produced `aedat` files, which were necessary to handle the combined event, image, and IMU data from these sensors. Later versions of the format also offer good compression for the data. Accessing the data is somewhat complicated by multiple versions of the file format, and also multiple libraries and tools capable of programmatically loading the data. These tools are not always comparable or consistent in how they load data, especially in terms of backward compatibility. The format also supports advanced features such as the interleaving of multiple camera streams, as in DHP19 [12], in which four different event camera streams are multiplexed into a single file. Whilst this is technically allowed and specified in the file format, very few of the tools and libraries support these features. However, the `aedat` format benefits greatly from the jAER tool [7], which allow browsing and exploring the data with a GUI interface.

By contrast, there is no dedicated category for the output of the Prophesee sensor. The Prophesee sensors, and their predecessors, the Chronocam and ATIS cameras, used a variety of different formats, including hand-rolled binary files. These converged into the better-defined and versioned EVT formats[3], but it remains difficult to determine which file formats are used in each dataset. For simplicity and clarity, these formats were labelled as `Binary`, with the Prophesee sensor data comprising the bulk of the datasets in that category. By contrast, the `aedat` format is a self-describing, versioned format with a file extension convention and backward-compatible tools for loading and working with the data, warranting a dedicated category for it.

---

[3]EVT files still contain plain-text comments at the beginning of each file that contain important information, such as sensor type and resolution. The structure of this metadata is not specified and has changed over time, making it difficult to automatically extract this information.

[4]The definition of the format can be found at https://github.com/neuromorphic-paris/event_stream.

The `EventStream`[4] format set out to be a clearly defined and open specification for event-based data. It supports light compression of the timestamps, and has a wide range of open-source tools to support it. However, it does have an express means for handling metadata, other data streams, and ground truth annotations.

The `Numpy` file format is perhaps the fastest-growing format, as shown in Fig. 4 where it is currently the most common format by dataset count. It is also rapidly catching up in terms of the total data volume. As a format, it is not specifically designed for neuromorphic data, but rather the default method for storing data from the `Numpy` scientific computing Python library. It supports compression and is a very efficient method to load and use data in Python. It can save multiple variables into a single file, leveraging the features of the python library to support self-describing structured arrays. The named columns in these structured arrays solves the issue of identifying the ordering of the components of each event, but does introduce a related problem in how each of the data types (i.e. `dtypes`) for each column is defined. This can make data from different groups incompatible, or even induce unexpected rounding and data truncation errors.

### B. The move to open standards

There is a clear trend toward more open data formats in recent years. The community has shifted from proprietary binary formats, to structured data formats, and finally are settling on open-source and simple file formats that offer ease of access, often at the cost of organisation and compression efficiency.

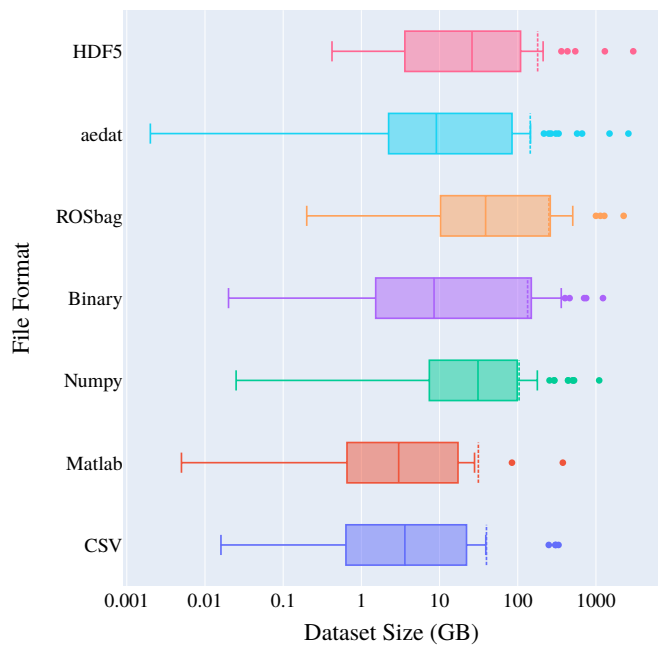

Fig. 5. **The distribution of dataset sizes by file format**. The above plot shows a plot of the distribution of dataset sizes for each of the formats. The mean dataset size is shown as the vertical line within each box. It is clear from the plot above that the `Matlab` and `CSV` file formats are typically used for smaller datasets, whilst the `ROSBag` and `Numpy` datasets are typically much larger in size.

Fig. 5 shows a box-and-whisker plot showing the spread of dataset sizes for each dataset format. The `Other` category has been omitted, as there is no consistency in the formats within that category. The mean dataset size for each format is shown as the vertical line within the box for each format.

Early neuromorphic datasets were almost exclusively distributed as binary files, often representing the raw data received from the sensor. This was common with datasets recorded using the ATIS camera, such as N-CARS [33] and the original EBBSSA [42], as there were no other file formats available at the time. Due to the maturity of jAER, data recorded with the DVS and DAVIS sensors, such as evbench [43] and DVSFLOW16 [44] were already using the `aedat` format. Both the `aedat` and `Binary` formats encompass a range of sensors, including the latest generation sensors with high spatial resolution, resulting in the wide spread of dataset sizes shown in Fig. 5.

Alongside the early growth of neuromorphic datasets was the use of Matlab as a tool to process event-based data, and many early datasets were released as Matlab data files. Some of the oldest datasets, such as MNIST-DVS [45] and DVS_Barrel [46] were distributed in this format. As is clear from Fig. 4, Matlab usage has largely plateaued, likely caused by the community shifting away from Matlab and towards Python for data processing. However, datasets are still occasionally released as Matlab data files, with noticeable examples including the wavefront sensing dataset EBWFNet [47] and the Event-based Bimodal Target dataset (EBT) [48].

The use of data formats from other research fields inevitably followed the rising interest in neuromorphic sensing, especially noticeable through the use of `ROSbags`. Largely driven by the robotics community, `ROSBags` started gaining popularity as means of distributing multi-modal datasets. Many of these datasets were from moving vehicles, resulting in large volumes of event-based data and making this category of dataset some of the largest in the community. Similarly, the `HDF5` format saw adoption from multi-modal datasets, such as 3ET [49] and 3ET+ [50], in which frames and event data are stored alongside one another. This is clearly evident in the breakdown of data volume by file format, shown in Fig. 4, in which almost 40% of all neuromorphic data is stored in `HDF5` and `ROSBags`.

Early datasets, such as DDD17 [51] in 2017 and MVSEC [28] from 2018 ushered in a wave of large multi-modal vehicle datasets, a trend that continues to this day through datasets such as ADD [36], DDD20 [52], the 1Mpx Detection Dataset [53] in 2020, the M2P2 [54] and the MA-VIED [55] datasets in 2024. The majority of these datasets are distributed as either `ROSBag` or `HDF5` files, heavily making use of their structured format to store multiple datasets. The outliers for these formats, shown in Fig. 5, are almost exclusively driving-related datasets.

Text-based file formats, such as `CSV`, have always been used to distribute event data, starting with the RGB-D [56] in 2014. It represented the most accessible data format, but also the most simple. It remains a popular means of distributing data, but only makes up a small percentage of the total volume of neuromorphic data. This is also evident in the spread of dataset sizes in Fig. 5, where the `CSV` format has one of the lowest mean sizes and the least significant outliers.

The use of `Numpy` as a file format only started in 2020 with the DVS-EMG dataset [57], which has become the most widely used dataset in that format. `Numpy` quickly grew in popularity, to the extent that was used in over 20% of all datasets in 2025. `Numpy` is the defacto Python library for scientific and mathematical computing and natively supporting compression. It also provides structured data types that lend themselves well to storing event-based data. The widespread adoption of Python for deep learning tool chains has enabled `Numpy` datasets to be easily integrated into deep and convolutional neural networks. The format supports both small datasets, such as the 1 GB THU-HSEVI dataset [58] and the 2 GB FlashyBackdoor [59], and very large datasets that include the 400 GB N-ImageNet [26] and the colossal 1.1 TB N-EPIC-Kitchens dataset [60].

*C. The issue of time*

Perhaps the most touted benefit of neuromorphic event-based data is the high temporal resolution, which manifests itself in the timestamps applied to each event or spike. In a typical event-based camera recording, over 60% of the bits used to store each event is dedicated to the timestamp. It would seem logical that the timing used for event-based data would have a well-defined convention and standard, but this is not the case.

Neuromorphic vision sensors typically produce timestamps with microsecond resolution, which is generated using a 1 MHz clock signal and counter within the device hardware. Whenever an event is generated, the value of the counter is captured and used as the timestamp for that event. The counter value therefore provides only relative timing information, as the counter will often continue to increase as long as the device is powered.

Recordings can therefore start at an arbitrary timestamp value, with the accurate timestamps increasing relative to that starting value. However, datasets sometimes subtract this value from all subsequent events to ensure that all recordings start at $t = 0$. This can become an issue when datasets involve synchronisation or time alignment between recordings, where the starting timestamp has significance. This can also cause issues when combining data, where the timestamps need to be aligned, as simultaneous recordings from different cameras may have different starting timestamps. These facets are almost never documented and can lead to subtle issues and errors that propagate through entire systems.

A second issue is how time is represented within each event. The most common approach stores the timestamp value as an unsigned integer, representing milliseconds. However, some datasets choose to store a different representation of time. For example, the Event-ReID [61] dataset only stores millisecond time resolution, whilst DVS-Lip [62] uses a signed integer to store the timestamps, allowing for potential negative timestamps.

Other datasets, such as The Event Camera Dataset [30] store the time in the `ROSTime` format, despite using `Numpy` to distribute the data. The EventGAN dataset [63] also uses

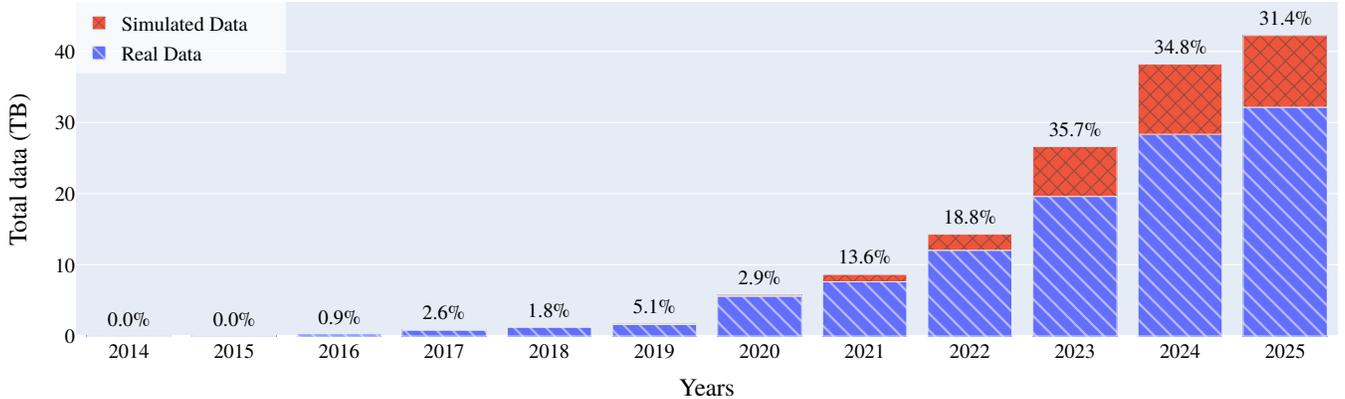

Fig. 6. **The growth of simulated event-based datasets.** The above figure shows the cumulative amount of neuromorphic data available each year, broken down into the amount of real data and the amount of simulated event-based data. The value above each column shows the percentage of total data that is simulated.

`ROSTime`, storing the values in an `HDF5` file. These issues at best create confusion and at worst introduce subtle errors into algorithms and systems.

### D. The issue of spatial precision

Event-based vision data does not inherently imply the resolution of the data. Unlike conventional imaging, where the frame size provides the spatial resolution, it is not always possible to determine the range of the $x$ and $y$ values from the event stream alone. Datasets, such as the 3ET [49] crop the event stream to a smaller spatial resolution, whilst simulated datasets can often pick spatial resolutions that match their source datasets. Simulated datasets, often maintain the spatial resolution of the source dataset from which they were converted, with N-Caltech101 further complicating the issue by having different resolutions across the input classes.

The output of an event-based vision sensor always includes the spatial location of the pixel emitting the event, such that $\boldsymbol{u} = (x, y)$, where $x$ and $y$ are integers indicating the row and column of the pixel causing the change. In most event-based data formats, these values are fittingly stored as unsigned integers.

However, datasets that involve coordinate transformations or homography, occasionally convert the pixel locations and report the locations in a different coordinate system, such as the SM-ERGB dataset [64], in which sub-pixel locations are provided. This is also common in space imaging datasets, where the positions of the events are converted from camera coordinates to sky coordinates. Such conversions inherently lead to a loss of information through the imprecision of the conversion from integer values to floating point value. Reversing this transformation can lead to values mapped incorrect back to camera pixels. Whilst it might appear good sense to preprocess the data into a more readily applicable coordinate system, it can severely limit the wider use of that data.

## V. The Rise of Simulated Event-based Data

Simulating event-based vision data is a very tempting proposition, albeit one that needs to be approached with caution and a careful understanding of the associated limitations and pitfalls.

TABLE III
Categorisation of real and simulated datasets

| | **Dataset Type** | **Example Datasets** |
|---|---|---|
| **Simulated Data** | **End-to-end simulation** Data sourced from ray-tracing or digital simulation. Events are generated with a simulator (e.g. ESIM). | CED |
| | **Video-to-Events** Data sourced from frames captured with a physical conventional camera and then converted using a event camera simulator (e.g. v2e). | DND21 |
| **Quasi-Real Data** | **Monitor Conversions** Data is sourced from static and dynamic scenes shown on a monitor, screen, or projector. A physical event camera is used to capture the events. | N-MNIST |
| **Real Data** | **Scene recordings** Data is sourced from real-world scenes. The events are recorded using a physical event camera. | DVS-Gesture |

There has been significant growth in the percentage of available data generated through means other than a physical sensor. Fig. 6 shows a plot of the cumulative amount of neuromorphic data over time, with each bar showing the relative amount of real and simulated data contributing to the total available data. Above each bar, the percentage of the total data that is simulated is provided. The graph provides insight into the rise, and start of a potential decline, in the use of simulated data.

The graph shows a very distinct upswing in the generation of simulated data in 2020, with the percentage of simulated data increasing by over 10% in that year alone. This growth continued until 2024, with a significant jump in 2023, likely corresponding to the rapid growth in the use of deep learning techniques which necessitate large volumes of data.

Interestingly, this trend seems to have slowed in the past two years, with a small decrease in the overall percentage of simulated data in 2024, potentially carrying across into 2025. This might be indicative of a shift away from simulated data, perhaps arising from the widespread availability of the current generation of neuromorphic sensors.

Exploring the impacts of this trend require a deep understanding of the means by which simulated data is produced. Table III provides a breakdown of the different event generation methods and their categorisation into simulated or real data for the context of this paper. Event-based vision data can be simulated by using software tools to convert existing frame-based video into an event stream using tools like `v2e` [65], through ray-tracing simulation in a 3D environment using a tool like `ESIM` [66], or through a hybrid approach in which a physical sensor is used to capture data from a monitor or screen, as used in N-MNIST [29] and DVS-MNIST [45].

However, this third method of conversion is somewhat controversial. It is typically referred to as a monitor conversion and this means of generating data remains an ongoing source of contention in the neuromorphic community. Initially, these datasets were created by flashing or moving images across an LCD monitor, as used in FLASH-MNIST [67], which resulted in concerns from the community over flickering artefacts caused by the screen refreshing.

N-MNIST and N-Caltech101 [29] attempted to mitigate these effects through keeping a static image on a fully illuminated LCD screen and moving the sensor, which reduced the flicker, but left critics sceptical of the temporal benefits of the data [68]. More recent attempts have tried to further mitigate these issues through the use of static e-ink screens, as in the SpiReco [69] dataset.

However, there are still concerns over the overly-sanitised nature of the data, and the lack of real-world noise and motion effects. Yet, monitor conversions continue to be popular, with recent datasets such as Unresolved Point-Object Centroiding Dataset (UPOC) [70] and the N-ImageNet dataset [26]. For the purposes of this work, these datasets are considered to be real dataset, as they are captured with a physical neuromorphic sensor, however they are all tagged as "Monitor Conversions" to make the distinction clear.

### A. The benefit of simulated data

There are compelling reasons for the use of simulated event-based datasets. Simulation enables the creation of data that would be challenging to collect with an event-based sensor, due to prohibitive costs, practical considerations, and even safety or regulatory restrictions. It also enables the use of existing datasets, providing access to the enormous corpus of existing data collected for the deep learning community and allowing for the exploration of applications that may be prohibitively expensive or impossible to directly explore, such as lunar navigation as in the Synthetic Lunar Terrain (SLT) dataset [71] or for automotive vehicle crashes as in the Event-Based Automotive Collision Detection Dataset (EBACDD) [72].

Perhaps most importantly, the use of simulated data removes some of the barriers to entry for the field, lowering the cost and complexity of acquiring data with a physical event-based sensor. These barriers have lessened in severity over the past ten years, due to the decreasing cost of neuromorphic sensors and the rise of existing datasets, but the benefit of low-cost and easy access to data remains of paramount importance to the growth and sustainability of the field.

### B. The pitfalls of simulated data

The core issues with simulation arise from our current knowledge of the sensors and our limited ability to accurately predict their performance in novel applications.

When converting video data into events, the output produced is inherently limited by the nature of the frame-based sensor capturing the source data. This can be significantly different from what a physical neuromorphic sensor would capture. Whilst the data from an event-based sensor resembles the output of a frame-differencing algorithm, it is subtly but fundamentally different, arising from the pixel circuitry itself.

For example, event-based sensors are not constrained by the same relationship between exposure time and frame rate, allowing high-speed phenomena to be captured at light levels that would be impractical for a conventional camera. This has been aptly demonstrated with lightning [73] and sprite [74] observations. Conversions from conventional datasets for these applications mask the subtle but important differences in what the sensor can capture.

End-to-end simulation approaches suffer from a similar issue, stemming from the fact that we can only simulate based on information that we know. This issue is two-fold, as there are currently no consistent models for event-based sensor pixels, and the elements of most simulations are often drawn upon information gathered from conventional sensors.

Whilst simulated data can pose significant concerns, it is not without merit, and certainly plays an important role in the research field. Simulated data is extremely beneficial when the underlying application is well understood and when the validity of the data can be determined or verified. Simulated data should not be used to explore new applications, as the data provided will always be limited by either our current understanding of the problem, or by the limitation of the existing sensors used to validate the simulation.

## VI. The Lack of Implied Context

A major issues that affects neuromorphic datasets is the inherent lack of context in event-based data, especially when compared to similar datasets in conventional image processing and deep learning.

Image datasets can trace their roots back to film-based cameras, whose original purpose was to capture visual representations that humans could readily understand and process. One can simply look at an image from a dataset, or frames from a video dataset, and easily understand and parse the visual scene. This parsing and understanding is rather extensive. Without any additional information, the subject or point of interest in the dataset can often be determined, along with the framing and the scene structure, the nature of the background and lighting conditions, the angle of the sensor relative to the scene, and often the entire purpose of the dataset.

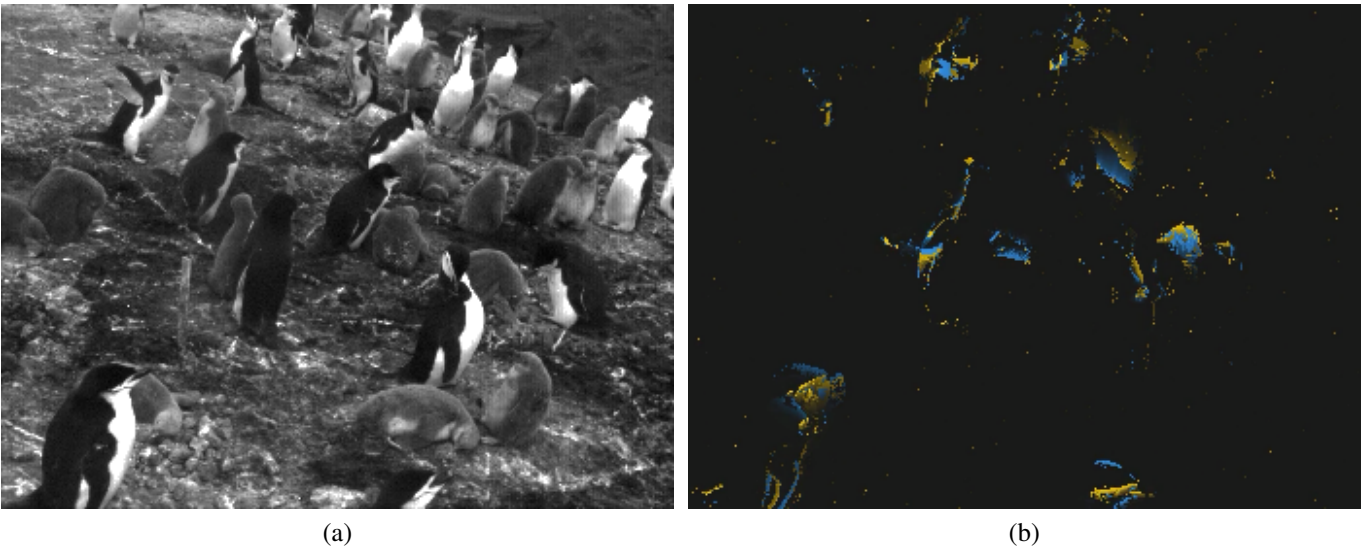

(a)                  (b)

Fig. 7. **Illustration of the inherent lack of context in event-based data.** The above image shows two representations of the same scene captured from a DAVIS346 sensor. The scene can be easily determined by simply looking at the raw image data shown in (a), whereas it is unlikely that one could determine the scene from the frame-based representation of the event-based data shown in (b). Both frame and event-based representation were sourced from the dataset. Data sourced from the dataset in [75].

This is not the case for neuromorphic data. Fig. 7 shows data from a real-world and publicly available neuromorphic dataset captured with a DAVIS346 sensor, a that sensor can capture both frames and events simultaneously. This allows for direct comparison of the same scene from a conventional image (left) and a similar visual representation of event-based data (right).

Although it may take a few seconds to visually assess the scene in (a), the subject and context can be determined without any clues to the nature of the dataset. This is not the case for the event-based data shown in (b).

It is worth noting that the data in (b) is a visualisation of the underlying event-based data. However, there is currently no repeatable or standard means of visually representing event-based data in an easily parsable format. One could potentially create a different 2D representation, or even a 3D spatiotemporal plot, that may prove to be more illustrative of the underlying data, but therein lies the point - neuromorphic data needs either special care when presenting data or clear and concise metadata and descriptions for it to be useful beyond the implementation for which it was collected.

For the visual representation of the dataset in (b), an accumulated frame from the Event Penguins [75] was created using a time duration set to match the frame rate of the conventional sensor. This provided the clearest view of the scene, with longer or shorter time durations making it even more difficult to match to the frame, let alone to glean any information from the data.

Taken to the extreme, a dataset consisting of objects that rarely move would provide neuromorphic data that consists of little other than sensor noise. A frame-based dataset variant would still likely provide enough information to understand the task, whereas the event-based data would be indistinguishable from background noise. For this case, it is the redundant information in the scene that provides the context for the frame-based image, which just happens to be what the event-based sensor removes.

This should be viewed as feature, rather than an issue with neuromorphic sensors. It is precisely this suppression of redundant data that provides the event-based sensor with its unique capabilities, but this does come at the cost of accessibility. Neuromorphic datasets therefore need to provide clear, concise, and detailed descriptions of their contained data to extol the virtues of the sensor and yet remain accessible for the research community.

## VII. Conclusions

Analysing the data collected on the available neuromorphic datasets shows some clear issues faced by the neuromorphic community. The use of citations as a proxy for use revealed two problems. Firstly, that most papers cite only one dataset which limits the applicability, ambition, and validity of many algorithms and studies. Secondly, that most papers also cite the same small subset of papers.

This leads to an important question - with whom does the fault lie for this situation? Is it with the paper authors? The analysis of the reusability of datasets discussed in Section II shows troubling issues with the way that researchers approach data in the neuromorphic community. For example, the rapid growth in the number of datasets coupled with the low median citation rate across datasets lends credence to the reality that authors often create new datasets rather than use existing data and also tend not to validate their work on multiple datasets.

However, datasets are at least partially to blame. In particular, datasets are often not available (the issue of data availability discussed in Section III), using multiple datasets is technically challenging (the issue of data accessibility discussed in Section IV), the suitability of a dataset is often difficult to ascertain (the lack of implied context discussed in Section VI), and even whether the datasets accurately represent the task being undertaken (the issue of simulated data discussed in Section V).

The purpose of this research is not only to serve as a means to creating successful datasets, but also as a guide to ensure that datasets are so easy to use that algorithm designers are enabled to test their ideas on multiple datasets, thereby bettering neuromorphic engineering. For this reason, this paper has focused on the trends in neuromorphic dataset production, generation, and usage over past ten years.

Based upon the findings in the sections above, we can draw conclusions around the best practices for datasets going forward. In this section, we distil the findings of this work into a set of recommendations for the community around the creation and usage of neuromorphic data.

*A. Reduce, Re-use, Re-process*

The growth in the quantity of available neuromorphic datasets is both encouraging and potentially concerning. As shown in Section II, there is a worrying trend in which new datasets are created instead of using existing datasets. Some of this can be explained through improvements in sensor technologies, necessitating the recreation or recapturing of similar datasets, but often the datasets are captured to cater toward a specific algorithm or approach.

Making use of existing datasets is fundamental to their existence, reducing the work required to support algorithm development and also to provide a means of benchmarking, comparing, and qualifying results and performance. This work is an effort to address this problem through highlighting best practices for dataset generation and distribution, suggesting mechanisms for re-using datasets more effectively, and by providing a tool to assist in dataset discovery. In support of these efforts, we suggest that the community consider the following suggestions:

**Focus on using existing datasets.** Whenever possible, use existing datasets instead of creating new datasets as it allows for better understanding and comparisons of the results, especially to existing work. Consider using datasets that were not directly intended for your desired application when they contain the correct data. Before embarking on dataset creation, assess why the existing datasets are not suitable, and include this reasoning when publishing research and datasets.

**Amend and extend instead of replace.** When existing datasets are not fit for purpose, consider either amending them with new annotations or extending them with new recordings instead of creating entirely new datasets. There have been several highly successful neuromorphic datasets adopting this approach, which has the added benefit of adopting an existing group of users familiar with the data.

**Consider the meta-dataset approach.** There is merit in building new datasets from existing data, especially when using datasets not expressly collected for the purposes of the task being assessed. This can remove any bias in the experimental setup, and better demonstrate the robustness of an algorithm or system. Whilst constructing meta-datasets may be more work than simply re-using an existing dataset, the benefits to both the research and the community can be significant.

*B. Distribute data sustainably*

The analysis in Section III of data distribution methods indicates that a successful dataset should make the process of accessing and downloading the data as simple as possible. Datasets should also ensure that their contents are well documented, allowing users to understand the structure of the dataset prior to downloading it. Therefore, we can make the following recommendations for distributing neuromorphic datasets:

**Consider the long-term accessibility of the data.** Dataset longevity should not be tied to a specific person or institution and the data and its description should be stored in a long-term and persistent manner.

**Make the structure of the dataset available and clear.** Datasets should be structured for easy perusal and download. Monolithic compressed archives without detailed file structures should be avoided. If they cannot be avoided, provide an easy-to-access description of their contents.

**Simplify access to the data.** Datasets should be easy to access. Avoid using multi-step or interactive license terms or forms to request access. Consider using included license files in the dataset over requests for access whenever possible.

**Support automation.** The data shows that the most widely used datasets are the ones that are easy-to-access and that support automated downloading.

*C. Prioritise data accessibility*

The findings from Section IV show a distinct trend toward more open and accessible data formats for storing neuromorphic datasets. The prevailing dataset formats offer backwards-compatibility, stability, and ease-of-use through either self-describing formats or well-defined and well-supported format specifications. Based on our findings, we can suggest the following best practices for distributing data:

**Always pick accessible formats over compressed formats.** Despite the enormous growth in the size of current datasets, there has been an even greater increase in the availability of storage, memory, and bandwidth. Unless there is a compelling reason requiring a small dataset, always prioritise ease-of-use over size.

**Adopt standard formats wherever possible.** It is clear that the most widely used datasets rely heavily on standard and open-source dataset formats that have a widely supported ecosystem and mature tool chains. Use these whenever possible, keeping in mind that these formats are agnostic to the data stored within them, thus requiring an accompanying description of how the data is stored within those formats.

**Make no assumptions in terms of data formats.** There are no widely-accepted conventions when storing event-based or neuromorphic data. There are tightly integrated communities that have adopted their own standards, giving the illusion of a community standard, and these often lead to assumed knowledge that can make data inaccessible to others in the field. Describe the order in which $[x, y, t, p]$ are packed into columns, explicitly state whether time is relative or absolute, and whether $t = 0$ has any semantic meaning.

**Distribute raw events over processed data.** Always try to provide raw data before converting it to different spatial or temporal reference frames. Instead, provide scripts that apply the transformation or conversion, either in real-time during processing or as a pre-processing step for your system. Even simple transformations can introduce errors that cannot easily be reversed.

**Separate out co-collected data and ground truth.** Co-collected data and associated ground truth are crucially important for specific applications, but can limit the applicability of data for other uses. Whenever possible, this accompanying data should be stored alongside the raw events, and linked through mappings rather than through modifications to raw event data.

### D. Remember, simulate responsibly

Simulated data is a double-edged sword. As discussed in Section V, it can greatly bolster the volume of available training data, but it can also be very misleading when exploring new applications. Care needs to be taken when using simulated data, and we can recommend the following approaches when dealing with simulated data:

**Simulate the known. Capture the unknown.** Perhaps the most important recommendation is to use simulation when the nature of the task is known. Use extreme caution when using simulators or converters to explore the use of neuromorphic sensors for new and novel applications. Caveat such work appropriately to ensure that future researchers are not discouraged.

**Consider the source of the data.** When converting conventional data to event-based datasets, always consider the device or sensor used to capture the source data. Characteristics such as frame rate, dynamic range, and integration time should be considered for their limiting effects on the target application. When converting data, always provide as much information as possible on the nature of the source data and how it was acquired.

**Validate simulated data whenever possible.** Ensure that simulated data is verified against real-world data whenever this is possible. This can be done by collecting a small dataset with a physical sensor - a trend that is often done in existing datasets with the data used to verify algorithms trained on the simulated data. However, few of these works compare the nature of their real and simulated data.

**Pay attention to noise.** Current neuromorphic sensor simulators and video converters do not simulate noise in a realistic manner, which can lead to algorithms and approaches that perform poorly when transitioned to real-world sensors, especially as noise performance varies wildly across different sensors, even from the same generation.

### E. Describe your data

If there is one message to take away from this work, it is the importance of describing your data thoroughly and exhaustively. Section VI discusses the degree to which conventional datasets provide contextual information that is not present in most neuromorphic datasets. For this reason, we suggest the following guidelines for distributing datasets with context:

**Describe the environment.** The nature of neuromorphic vision sensors removes redundant information, which often contains contextual clues for the visual scene. Describe the geometry of the scene, the position of the sensor, the background of the target, the lighting conditions, and the nature of the scene itself. Whilst photographs can capture much of this information, keep in mind that the scene may be dynamic, and the nature of those changes need to be captured as well.

**Describe the camera motion.** Ensure that the nature of the camera and any camera movement is well described and documented. Camera movement is much harder to infer from neuromorphic data than conventional video data, making it necessary to explicitly describe the motion of the sensor in the scene. Scene understanding often requires knowledge of the environment in which the sensor is placed, making it seem unnecessary to those familiar with the task. However, without that context, the data is often difficult to understand or parse.

**Describe the task in detail.** It is crucial to describe the intended task in detail when distributing neuromorphic datasets. Describe the intended targets, the expected number of objects in the scene, what is expected to be moving, and what is expected to be stationary. Explain the rationale for choices such as camera positioning and motion, and highlight the temporal and spatial expectations for the data to be captured.

## VIII. The List of Available Neuromorphic datasets (LAND)

We have made the bulk of the data used in this work publicly available as an interactive tool for dataset discovery. Called the List of Available Neuromorphic datasets (LAND), this tool contains almost all the collected information for the 423 datasets used in this work, with the exception of some proprietary reference and citation data used in the generation of figures. LAND is intended to be a living tool, with the exact version representing the work in this paper available online[5]. The latest version of LAND can always be found online at https://neuromorphicsystems.github.io/land.

---

[5]The 2025 snapshot of the LAND dataset is available online at https://github.com/neuromorphicsystems/land/releases/tag/2025-Archive/.